\documentclass[pdflatex,sn-mathphys-num]{sn-jnl}

\usepackage[T1]{fontenc}
\usepackage[utf8]{inputenc}
\usepackage{graphicx}%
\usepackage{multirow}%
\usepackage{amsmath,amssymb,amsfonts}%
\usepackage{amsthm}%
\usepackage{mathrsfs}%
\usepackage{booktabs}   
\usepackage{array}      
\usepackage{placeins}
\usepackage[title]{appendix}%
\usepackage{xcolor}%
\usepackage{textcomp}%
\usepackage{manyfoot}%
\usepackage{booktabs}%
\usepackage{algorithm}%
\usepackage{algorithmicx}%
\usepackage{algpseudocode}%
\usepackage{listings}%
\usepackage{xcolor} 
\usepackage{booktabs}
\usepackage{tabularx}
\usepackage{multirow}
\usepackage{makecell}
\usepackage{array}
\usepackage{threeparttable}
\usepackage{subcaption}
\usepackage[utf8]{inputenc}   
\usepackage{subcaption}
\usepackage{caption}
\usepackage{url}
\usepackage{hyperref}

\theoremstyle{thmstyleone}%
\theoremstyle{thmstyletwo}%

\theoremstyle{thmstylethree}%

\begin{document}

\title[Article Title]{Silhouette-Based Gaitomics for Frailty Assessment in Older Adults}


\author[1]{\fnm{Laura} \sur{McDaniel}}\email{lmcdan11@jhu.edu}

\author[2,3]{\fnm{Ryan} \sur{Roemmich}}\email{rroemmi1@jhmi.edu}

\author[4]{\fnm{Crystal} \sur{Szczesny}}\email{cszczes1@jh.edu}

\author[4]{\fnm{Peter} \sur{Abadir}}\email{pabadir1@jhmi.edu}

\author[1,5]{\fnm{Rama} \sur{Chellappa}}\email{rchella4@jhu.edu}

\author[1]{\fnm{Basudha} \sur{Pal}}\email{bpal5@jhu.edu}


\affil[1]{\orgdiv{Department of Electrical and Computer Engineering}, 
\orgname{Johns Hopkins University}, 
\orgaddress{\city{Baltimore}, \state{MD}, \country{USA}}}

\affil[2]{\orgdiv{Center for Movement Studies}, 
\orgname{Kennedy Krieger Institute}, 
\orgaddress{\city{Baltimore}, \state{MD}, \country{USA}}}

\affil[3]{\orgdiv{Department of Physical Medicine and Rehabilitation}, 
\orgname{Johns Hopkins University School of Medicine}, 
\orgaddress{\city{Baltimore}, \state{MD}, \country{USA}}}

\affil[4]{\orgdiv{Division of Geriatrics and Gerontology Medicine}, 
\orgname{Johns Hopkins University School of Medicine}, 
\orgaddress{\city{Baltimore}, \state{MD}, \country{USA}}}

\affil[5]{\orgdiv{Department of Biomedical Engineering}, 
\orgname{Johns Hopkins University}, 
\orgaddress{\city{Baltimore}, \state{MD}, \country{USA}}}


\abstract{Walking offers a window into physical function that may support interpretable video-based frailty assessment. We developed silhouette-based ``gaitomics'' to connect frailty predictions with measurable patterns of body configuration and movement. In 66 older adults assessed using the Fried Frailty Phenotype, standardized silhouettes and gait energy images yielded 90 descriptors of shape, temporal variation, motion, and texture. Minimum redundancy maximum relevance selection using the mutual information difference criterion prioritized informative features while limiting redundancy. Each classifier was evaluated using nested stratified cross-validation with five outer and three inner folds at the participant level, with feature ranking and feature-count selection confined to training partitions. Shapley additive explanations examined contributions to held-out predictions. Bilateral silhouette difference led the aggregate feature-selection rankings for three-class classification and both comparisons involving frail participants. Attribution profiles differed across tasks, with texture and consecutive-frame silhouette differences prominent in three-class classification. Among classifier-specific evaluations, CatBoost achieved the highest mean three-class micro-AUC of \(0.647\pm0.085\). Leading pairwise mean micro-AUCs were \(0.833\pm0.249\) for frail versus nonfrail, \(0.765\pm0.168\) for frail versus prefrail, and \(0.554\pm0.129\) for prefrail versus nonfrail classification. These findings connect model predictions to explicit gait characteristics while revealing task-dependent feature priorities and limited separation of prefrail and nonfrail participants. The framework provides an interpretable basis for further investigation, with larger independent studies needed to establish reliability and clinical relevance.}




\maketitle

\section{Introduction}\label{sec1}

Frailty is a dynamic geriatric syndrome characterized by diminished physiological reserve and increased vulnerability to stressors, increasing the risk of adverse health outcomes and reduced quality of life~\cite{xue2011frailty}. It is associated with falls, fractures, impaired recovery, increased healthcare utilization and costs, limitations in daily activities, loss of independence, and mortality~\cite{sirven2017cost,zhang2025correlation,dent2024frailty,kim2024frailty,deng2025frailty}. Identifying and characterizing frailty is therefore important to preserving health and function in aging populations.
Population aging further emphasizes this need. The World Health Organization projects that one in six people globally will be aged 60 years or older by 2030. By 2050, this population is expected to exceed twice its 2020 size, reaching approximately 2.1 billion, while the population aged 80 years or older is expected to triple to approximately 426 million~\cite{rudnicka2020world}. The United Nations projects that adults aged 65 years or older will outnumber children younger than 18 years by the late 2070s~\cite{united2022peace}. Frailty prevalence also increases with age, with reported estimates among community-dwelling individuals ranging from approximately 11\% at ages 50--59 years to 51\% at ages 90 years or older~\cite{o2021prevalence}. These trends underscore the need for scalable, objective approaches that complement clinical assessment.

More than seventy frailty assessment instruments have been described, including questionnaires, multidimensional indices, and physical performance assessments, yet none is optimal across all screening settings~\cite{deng2024global}. The Fried Frailty Phenotype is among the most widely established measures of physical frailty~\cite{fried2001frailty,le2022development}. However, assessments may depend on subjective information, dedicated administration time, specialized equipment, trained personnel, or controlled environments~\cite{wang2024scale}. These requirements can constrain frequent or large-scale assessment, motivating complementary digital measurements that can be acquired and interpreted consistently.
Gait provides an observable marker of physical function because walking depends on coordinated musculoskeletal, neurological, sensory, and cardiovascular activity. Changes in gait may therefore reflect multisystem decline and reduced functional reserve~\cite{taylor2023multisystem,studenski2011gait}. Advances in computer vision and machine learning have enabled quantitative video-based analysis, with machine learning and deep learning models showing promise for identifying frailty-related information~\cite{mcdaniel2026gait,wang2025combo}.
Silhouette sequences retain body geometry and movement while reducing facial, clothing, and environmental detail visible in raw video. They provide a practical representation for analyzing shape and motion~\cite{padilla2015visual,collins2002silhouette}, supporting privacy-conscious analysis without guaranteeing anonymity. Their explicit boundaries and temporal structure also facilitate the extraction of measurements that can be examined alongside model predictions.
Interpretability remains a challenge for clinical translation because many predictive models do not directly communicate which measurable gait characteristics contribute to classification~\cite{yang2022unbox,hassija2024interpreting}. This matters when distinguishing nonfrail, prefrail, and frail individuals, since a predicted category alone offers limited insight into the observed movement underlying that prediction. Difficulty connecting outputs to explicit measurements may reduce clinical trust, impede examination of model behavior, and complicate clinical use~\cite{miotto2018deep,esteva2021deep,pal2025renal}. Quantitative movement descriptors can complement predictive performance by making the basis of classification more accessible.

Medical imaging offers a methodological precedent, with predictive applications facing similar questions about how image information contributes to clinical predictions~\cite{wang2025clinical}. Radiomics systematically extracts predefined shape, intensity, and texture measurements, converting images into structured data for analysis alongside clinical phenotypes and outcomes~\cite{mayerhoefer2020introduction}. Applications include tumor characterization, treatment response, and outcome prediction~\cite{chen2022application,gangil2022utility}, with extensions to cardiac, neurological, and musculoskeletal tissues. Explicit feature definitions support examination of model inputs, although connecting these measurements to model behavior and clinical meaning remains an active research challenge~\cite{xu2025addressing,ray2025review}.
Inspired by radiomics, we use the term \emph{gaitomics} to describe the systematic extraction, selection, and interpretation of quantitative descriptors from gait silhouettes and their accumulated image representations. This framework adapts the radiomics principle of a structured feature catalogue to body configuration and its spatial and temporal variation during walking. Geometry descriptors characterize projected shape, temporal descriptors summarize changes across frames, and motion descriptors quantify image-derived movement. Gait energy images represent accumulated silhouette occupancy, supporting intensity and texture measurements. Together, these descriptors characterize complementary aspects of gait that can be related to clinical assessments and model predictions.
This formulation supports investigation of informative measurements, redundancy among features, and their contributions to classification. It also provides an interpretable reference for future comparisons with learned gait representations. Feature-level analysis is particularly relevant to prefrailty, where less pronounced or heterogeneous differences may complicate discrimination from nonfrail individuals. The present study examines differences among contemporaneously assessed frailty categories.
We evaluate silhouette-based gaitomics in 66 older adults assessed using the Fried Frailty Phenotype. The study investigates whether explicit measurements of body configuration and movement support frailty classification and clarify the characteristics contributing to predictions. Its principal contributions are summarized below.

\begin{enumerate}
    \item We establish a structured catalogue of 90 handcrafted descriptors spanning silhouette shape, temporal variation, image-derived motion, and gait energy image intensity and texture. This catalogue provides a defined framework for quantifying complementary aspects of recorded gait.

    \item We evaluate feature subset size and predictive performance across five classifiers for three-class classification and each pairwise frailty comparison. Minimum redundancy maximum relevance selection using the mutual information difference criterion ranks features within the relevant training partitions for each task. Subsets are evaluated in increments of ten features using nested stratified cross-validation with five outer and three inner folds at the participant level. For each classifier, feature-count selection uses inner-validation performance, followed by evaluation on held-out outer-test participants. Classifiers are compared descriptively using their outer-test performance summaries.

    \item We characterize feature contributions using Shapley additive explanations from held-out outer-test participants. Visualizations display the ten highest-importance features while retaining the complete input set of each explained model. Examination of feature recurrence across aggregate outer-training rankings further identifies measurements shared across classification tasks.
\end{enumerate}

This framework connects quantitative gait characterization with predictive evaluation and post hoc interpretation, supporting further investigation of video-derived measurements as a complement to clinical frailty assessment.

\captionsetup[table]{labelsep=period}

\section{Methods}\label{sec11}

\subsection{Study Population, Gait Acquisition, and Video Processing}

The study included 66 older adults who were classified using the Fried Frailty Phenotype before gait acquisition. The cohort comprised 25 nonfrail, 24 prefrail, and 17 frail participants. The study was approved by the Johns Hopkins Institutional Review Board under approval number 00255175. All participants provided written informed consent before participation. Eligibility required the ability to walk continuously for five minutes without stopping. Exclusion criteria were neurological conditions, including Parkinson's disease, stroke, or cerebellar ataxia, congestive heart failure, peripheral artery disease with claudication, active cancer, pulmonary or renal failure, unstable angina, uncontrolled hypertension, dementia, severe aphasia, asthma that interfered with oxygen-consumption testing, and orthopedic or pain conditions that could compromise safe participation. Participants completed the walking task at a comfortable, self-selected pace and were permitted to use customary walking aids for safety. Demographic and clinical characteristics are summarized in Table~\ref{tab:demographics}.

\begin{table}[ht]
\centering
\caption{Demographic and clinical characteristics of the study cohort, stratified by Fried frailty status. Values are mean or $n$ (\%).}
\label{tab:demographics}
\begin{tabular}{lcccc}
\toprule
Characteristic & Nonfrail & Prefrail & Frail & Total \\
 & ($n=25$) & ($n=24$) & ($n=17$) & ($n=66$) \\
\midrule
Age, years & 77.4 & 79.7 & 79.8 & 78.8 \\
\addlinespace
Sex, $n$ (\%) & & & & \\
\quad Female & 11 (44.0) & 13 (54.2) & 15 (88.2) & 39 (59.1) \\
\quad Male & 14 (56.0) & 11 (45.8) & 2 (11.8) & 27 (40.9) \\
\addlinespace
Race, $n$ (\%) & & & & \\
\quad White & 23 (92.0) & 21 (87.5) & 11 (64.7) & 55 (83.3) \\
\quad African American & 1 (4.0) & 2 (8.3) & 6 (35.3) & 9 (13.6) \\
\quad Asian & 1 (4.0) & 1 (4.2) & 0 (0.0) & 2 (3.0) \\
\bottomrule
\end{tabular}
\end{table}

Participants walked back and forth along a fixed diagonal path measuring 4~m in each direction, corresponding to an 8~m out-and-back traversal. Gait was recorded using the Qualisys Motion Capture System, which comprised eight Arqus marker-based infrared cameras and eight Miqus markerless video cameras positioned around the room perimeter. The Miqus recordings were used for person detection and silhouette extraction. Person-detection and silhouette-generation rates were evaluated across all eight Miqus views to identify the most suitable view for analysis. Across views, person-detection rates ranged from 53.83\% to 96.23\%, and silhouette-generation rates ranged from 45.43\% to 99.73\%. The view with the highest person-detection and silhouette-generation rates was selected for subsequent processing. Videos from this view had frame dimensions of $1088\times1920$ pixels and were recorded at 60 frames per second. Turns were manually removed before segmentation.

Binary participant silhouettes were generated using Segment Anything Model 2.1 with the Hiera-B+ checkpoint~\cite{ravi2024sam2}. A bounding box was placed around the participant in one frame, and the model tracked the segmentation forward and backward through the recording. Participants who used walking aids retained them during acquisition. The segmentation prompt targeted the participant, and walking aids were excluded from the retained mask. All masks were manually inspected to confirm that the participant was fully captured. Frames were excluded when the silhouette contacted the image border and the resulting mask was incomplete. The segmentation output was stored as $64\times64$ binary silhouette frames, with between 216 and 1{,}890 frames available per walk.

The retained masks were binarized and cropped using a participant-specific bounding box that enclosed the silhouette across the retained sequence. The bounding box was expanded by 5\% of its width and height plus two pixels on each side. The cropped masks were resized to a width of 64 pixels and a height of 144 pixels using area interpolation without preserving the original aspect ratio. Uniform index sampling then selected 96 evenly spaced frames. All recordings contained more than 96 frames, so padding was unnecessary. Each participant was represented by a standardized 96-frame region-of-interest (ROI) sequence. Feature extraction was performed in this standardized image coordinate system. Consecutive frames in the subsequent analysis refer to neighboring positions in the resampled sequence, which need not be consecutive frames in the original 60-frame-per-second recording. Temporal and motion descriptors therefore characterized changes across the sampled sequence rather than physical displacement per second or explicitly segmented gait cycles.

For participant $i$, the gait energy image (GEI) was computed as the temporal mean of the standardized binary silhouettes according to Equation~\ref{eq:gei_methods}.
\begin{equation}
G_i(x,y)=\frac{1}{T}\sum_{t=1}^{T}S_{i,t}(x,y),\qquad T=96.
\label{eq:gei_methods}
\end{equation}
Here, $S_{i,t}(x,y)\in\{0,1\}$ is the silhouette value at pixel $(x,y)$ in sampled frame $t$. A GEI value of one indicates that the pixel was occupied in every sampled frame, while a value of zero indicates that it was never occupied. Intermediate values represent the proportion of frames containing foreground at that location. The GEI summarizes accumulated silhouette occupancy, while the ROI sequence preserves the order of frame-level shape and movement changes.

\subsection{Gaitomic Feature Catalogue and Rationale}

\subsubsection{Feature Families and Feature Order}

The feature catalogue was designed to represent complementary properties of projected body configuration and its change during walking. Shape features describe silhouette geometry, while temporal features summarize how scalar measurements and bilateral configuration vary across the sequence. Spatiotemporal features describe changes between sampled frames and include measures of local image change and motion direction. GEI intensity features summarize the distribution of accumulated silhouette occupancy. GEI texture features describe how neighboring occupancy values are arranged spatially. These families allow similar overall silhouette measurements to be distinguished by their temporal behavior or spatial organization. For example, two sequences can exhibit similar area variability while differing in the direction or location of visible movement.

The catalogue comprised 49 shape features, 14 temporal features, nine spatiotemporal features, six GEI first-order intensity features, and 12 GEI texture features, giving 90 descriptors per participant. This total resulted from the predefined measurements and their summary statistics rather than a statistically determined target feature count. Means describe typical values across a sequence, standard deviations describe variation, and minima and maxima capture the range of observed configurations. Relative-variation measures express variability in relation to the corresponding mean. The catalogue was specified before feature selection and classification. The complete identifiers and definitions are provided in the Supplementary Material.

Feature family and feature order served different purposes. Family identified the measurement source and the aspect of the silhouette being characterized. Order referred to the operational feature-construction taxonomy adopted in this study. The 72 first-order descriptors comprised the 49 shape features, 14 temporal features, six GEI intensity features, and three aggregate frame-difference features. These descriptors summarized individual geometric measurements, scalar trajectories, intensity distributions, or the magnitude of frame differences. The 18 second-order descriptors comprised six gradient- or optical-flow-based spatiotemporal features and 12 GLCM-based GEI texture features. These descriptors used spatial relationships or derived spatial and motion fields. The spatiotemporal family consequently contained features assigned to both orders. All 90 features were considered jointly as candidates for mRMR-MID selection. Feature-order assignments were used to describe the selected representation, rather than to define separate classification experiments.

\begin{table}[htbp]
\centering
\caption{Gaitomic feature families extracted from the standardized ROI sequence and GEI. The catalogue contains 72 first-order and 18 second-order descriptors under the study's operational taxonomy. ROI denotes the region of interest. GEI denotes the gait energy image. GLCM denotes the gray-level co-occurrence matrix. ASM denotes angular second moment. Complete feature identifiers are provided in the Supplementary Material.}
\label{tab:feature_families}
\footnotesize
\setlength{\tabcolsep}{3pt}
\renewcommand{\arraystretch}{1.2}
\begin{tabular}{@{}llc >{\raggedright\arraybackslash}p{0.43\textwidth}@{}}
\toprule
\textbf{Feature family} & \textbf{Source} & \textbf{Count} & \textbf{Description} \\
\midrule
Shape & ROI sequence & 49 & These first-order features describe contour geometry and its sequence-level variation using area, perimeter, solidity, compactness, aspect ratio, and Hu moments. \\
Temporal & ROI sequence & 14 & These first-order features describe image-derived center-of-mass position and motion, foreground-area variability, consecutive-mask overlap, sequence-derived velocity, and bilateral silhouette differences. \\
Spatiotemporal & ROI sequence & 9 & These features describe consecutive-frame change, spatial gradients of that change, and optical flow. Three are first-order and six are second-order. \\
GEI first-order & GEI & 6 & These first-order features summarize intensity mean, standard deviation, skewness, excess kurtosis, energy, and entropy. \\
GEI texture & GEI & 12 & These second-order features summarize GLCM contrast, dissimilarity, homogeneity, energy, ASM, and correlation using their means and standard deviations across orientations. \\
\midrule
\textbf{Total} & & \textbf{90} & The candidate catalogue combines direct silhouette measurements with derived spatial and motion descriptors. \\
\bottomrule
\end{tabular}
\end{table}

\subsubsection{Shape Descriptors}

Shape descriptors were extracted from the largest external contour in each standardized silhouette using OpenCV. In the following equations, the participant index is omitted for clarity, and $S_t$ denotes a standardized mask from one participant. For its largest external contour $C_t$, contour area $A_t$, perimeter $P_t$, solidity, compactness, and aspect ratio were computed as given in Equation~\ref{eq:contour}.
\begin{equation}
\begin{aligned}
A_t &= \operatorname{Area}(C_t), \qquad
P_t = \operatorname{Perimeter}(C_t),\\
\operatorname{Solidity}_t &= \frac{A_t}{A_t^{\mathrm{hull}}}, \qquad
A_t^{\mathrm{hull}} = \operatorname{Area}\bigl(\operatorname{ConvexHull}(C_t)\bigr),\\
\operatorname{Compactness}_t &= \frac{4\pi A_t}{P_t^{2}}, \qquad
\operatorname{Aspect}_t = \frac{w_t}{h_t}.
\end{aligned}
\label{eq:contour}
\end{equation}
Here, $w_t$ and $h_t$ are the width and height of the contour's axis-aligned bounding box. Area and perimeter describe silhouette extent and boundary length. Solidity describes the proportion of the convex hull occupied by the contour, while compactness relates enclosed area to boundary length. Aspect ratio describes the relative width and height of the observed contour within the standardized image.

Seven Hu moments described the spatial distribution of contour shape. They were computed from the contour moments and transformed using Equation~\ref{eq:hu} to compress their dynamic range. For consistency with feature identifiers such as \texttt{shape\_hu0\_mean}, the seven descriptors are indexed from zero to six in the notation below.
\begin{equation}
\begin{gathered}
H_{r,t} = \operatorname{HuMoments}\bigl(\operatorname{moments}(C_t)\bigr)_r,
\qquad r=0,\ldots,6,\\
\widetilde{H}_{r,t} = -\operatorname{sign}(H_{r,t})\log_{10}\bigl(|H_{r,t}|+10^{-12}\bigr).
\end{gathered}
\label{eq:hu}
\end{equation}
The five contour descriptors and seven transformed Hu moments gave 12 frame-level properties. Each was summarized across frames using its mean, standard deviation, minimum, and maximum, producing 48 features. The remaining shape feature, \texttt{shape\_frame\_area\_cv}, was the coefficient of variation of contour area across frames. Thus, the shape family retained both typical silhouette configuration and its variation over the recorded sequence.

\subsubsection{Temporal Descriptors}

The 14 temporal descriptors characterized image-derived center-of-mass position and motion, foreground-pixel area, overlap between consecutive masks, sequence-derived velocity, and bilateral silhouette configuration. Center-of-mass descriptors represented the position of the binary foreground within the image and were not direct measurements of the physical whole-body center of mass. Overlap descriptors summarized similarity between neighboring masks, while area and position summaries characterized the extent of sequence-level change. These measurements described the standardized sampled sequence rather than variability between individually identified steps.

Contour-based area and foreground-pixel area were retained as distinct measurements. The feature \texttt{shape\_area\_std} summarized the standard deviation of the area enclosed by the largest external contour. The feature \texttt{temp\_area\_std} summarized the standard deviation of the number of foreground pixels. Their relative-variation counterparts were \texttt{shape\_frame\_area\_cv} and \texttt{temp\_area\_cv}. This distinction preserved the different definitions of area used by the shape and temporal families.

The bilateral descriptor \texttt{temp\_sym\_absdiff\_mean} compared the portions of a silhouette on either side of its image-derived center-of-mass column. The two portions were padded to equal width, and the right portion was mirrored horizontally before comparison. The frame-level absolute difference was computed using Equation~\ref{eq:bilateral}.
\begin{equation}
s_t = \operatorname*{mean}_{x,y}
\left|
\widetilde{S}^{\mathrm{left}}_t(x,y)
-\operatorname{flip}_x\bigl(\widetilde{S}^{\mathrm{right}}_t(x,y)\bigr)
\right|.
\label{eq:bilateral}
\end{equation}
The feature was the mean of $s_t$ across the sampled frames. It quantified differences in projected bilateral silhouette configuration. It was not defined as a measurement of bilateral step timing, limb loading, or joint kinematics.

\subsubsection{Spatiotemporal Descriptors}

Spatiotemporal descriptors characterized changes between neighboring frames of the standardized sequence. Aggregate frame-difference features measured the amount of silhouette change, while spatial-gradient features described the spatial variation of those changes. Optical-flow features summarized estimated image motion and its direction.

For each consecutive pair, the foreground union $U_t$ and mean absolute frame difference $d_t$ were defined by Equation~\ref{eq:st_first}.
\begin{equation}
\begin{gathered}
U_t=\bigl\{(x,y)\mid\max\bigl(S_t(x,y),S_{t-1}(x,y)\bigr)>0.5\bigr\},\\
d_t=\frac{1}{|U_t|}\sum_{(x,y)\in U_t}
\bigl|S_t(x,y)-S_{t-1}(x,y)\bigr|.
\end{gathered}
\label{eq:st_first}
\end{equation}
The mean, standard deviation, and maximum of $d_t$ yielded the three first-order spatiotemporal features \texttt{st\_fdiff\_mean}, \texttt{st\_fdiff\_std}, and \texttt{st\_fdiff\_max}.

The absolute frame-difference image was also processed using horizontal and vertical Sobel operators with kernel size three. The resulting summary was computed according to Equation~\ref{eq:st_sobel}.
\begin{equation}
\Delta_t=|S_t-S_{t-1}|,\qquad
g_t=\operatorname*{mean}_{x,y}
\left(\bigl|G_x(\Delta_t)\bigr|+\bigl|G_y(\Delta_t)\bigr|\right).
\label{eq:st_sobel}
\end{equation}
Here, $G_x$ and $G_y$ denote the Sobel operators. The temporal mean and standard deviation of $g_t$ yielded \texttt{st\_fdt\_sobel\_mean} and \texttt{st\_fdt\_sobel\_std\_t}. These features described spatial changes in the difference image rather than only its overall magnitude.

Optical flow was estimated between neighboring standardized frames using OpenCV's Farneb\"ack method. The horizontal and vertical flow components, denoted by $f_x$ and $f_y$, were summarized over $U_t$ using Equations~\ref{eq:st_flow} and~\ref{eq:st_flow_ratio}.
\begin{equation}
\begin{gathered}
(f_x,f_y)=\operatorname{Farneb\ddot{a}ck}(S_{t-1},S_t),\\
m_t=\frac{1}{|U_t|}\sum_{(x,y)\in U_t}
\sqrt{f_x(x,y)^2+f_y(x,y)^2}.
\end{gathered}
\label{eq:st_flow}
\end{equation}
\begin{equation}
r_t=
\frac{\operatorname*{mean}_{(x,y)\in U_t}|f_y(x,y)|}
{\operatorname*{mean}_{(x,y)\in U_t}|f_x(x,y)|+10^{-8}}.
\label{eq:st_flow_ratio}
\end{equation}
The mean and standard deviation of $m_t$ yielded \texttt{st\_flow\_mag\_mean} and \texttt{st\_flow\_mag\_std}. The temporal mean of $r_t$ yielded \texttt{st\_flow\_vert\_ratio\_mean}. The temporal mean of the signed horizontal flow over $U_t$ yielded \texttt{st\_flow\_fx\_mean}. Together with the two Sobel summaries, these four optical-flow descriptors formed the six second-order spatiotemporal features.

\subsubsection{GEI Intensity and Texture Descriptors}

The six GEI intensity descriptors summarized the distribution of accumulated silhouette occupancy using the mean, standard deviation, skewness, excess kurtosis, energy, and entropy. The mean and standard deviation described the typical occupancy value and its spread. Skewness and excess kurtosis described additional properties of the intensity distribution. Energy summarized squared intensity, while entropy summarized the distribution of values across histogram bins. These quantities characterized the image representation and were not measurements of physical energy expenditure.

Descriptors were computed over GEI pixels with values greater than 0.01. The entire GEI was used when fewer than 50 pixels met this condition. For selected pixel values $x_j$ with mean $\bar{x}$ and standard deviation $\sigma$, the remaining intensity statistics were computed using Equation~\ref{eq:gei_first}.
\begin{equation}
\begin{gathered}
\operatorname{Skewness}=
\frac{\operatorname{mean}((x_j-\bar{x})^3)}{\sigma^3+10^{-12}},\\
\operatorname{Kurtosis}=
\frac{\operatorname{mean}((x_j-\bar{x})^4)}{\sigma^4+10^{-12}}-3,\\
\operatorname{Energy}=\frac{1}{N}\sum_{j=1}^{N}x_j^2,\qquad
\operatorname{Entropy}=-\sum_{k=1}^{32}p_k\ln p_k.
\end{gathered}
\label{eq:gei_first}
\end{equation}
Here, $N$ is the number of included pixels, and $p_k$ is the normalized probability of bin $k$ in a 32-bin histogram over $[0,1]$. Entropy used the natural logarithm, and kurtosis was expressed as excess kurtosis.

GEI texture features were derived from gray-level co-occurrence matrices, which describe how often pairs of intensity values occur at a specified spatial displacement. Unlike intensity-distribution summaries, these measurements retain information about relationships between neighboring pixels. GEI values were clipped to $[0,1]$ and quantized to 16 gray levels using Equation~\ref{eq:quant}.
\begin{equation}
q(x,y)=\operatorname{round}\bigl(15G_i(x,y)\bigr),\qquad
q(x,y)\in\{0,\ldots,15\}.
\label{eq:quant}
\end{equation}
GLCMs were computed at a displacement of one pixel in the four orientations $0^\circ$, $45^\circ$, $90^\circ$, and $135^\circ$. The matrices were symmetric and normalized. Symmetry meant that each pixel pair was counted in both directions. It did not involve mirroring or bilateral alignment of the GEI.

Six GLCM properties were evaluated. Contrast and dissimilarity described differences between paired gray levels, while homogeneity characterized the concentration of pairs with similar values. Angular second moment and its square root, termed GLCM energy, summarized concentration in the co-occurrence distribution. Correlation described the relationship between the paired gray levels. Each property was summarized using its mean and standard deviation across the four orientations, producing 12 texture features. GLCM energy was a texture statistic distinct from the mean squared GEI intensity used in the first-order catalogue.

\subsection{Feature Selection and Classification}

\subsubsection{Classification Tasks and Analysis Unit}

The participant was the unit of analysis. Each participant contributed one 90-feature vector and one reference frailty category. The three-class task used all 66 participants. Pairwise tasks compared frail with nonfrail participants using 42 participants, frail with prefrail participants using 41 participants, and prefrail with nonfrail participants using 49 participants. Features were ranked separately for each task within the relevant training partitions so that selection reflected the labels of that comparison without using its validation or test participants.

\subsubsection{mRMR-MID Feature Selection}

Features were ranked using minimum redundancy maximum relevance with the mutual information difference criterion, denoted by mRMR-MID~\cite{peng2005feature}. Mutual information measured dependence between a candidate feature and the task labels. The relevance term favored features informative about the labels, while the redundancy term penalized information shared with previously selected features. For candidate feature $X_j$, label variable $Y$, and the set $\mathcal{S}$ of features already selected, the MID score was defined by Equation~\ref{eq:mrmr_mid}.
\begin{equation}
\operatorname{MID}(X_j\mid\mathcal{S})=
I(X_j,Y)-\frac{1}{|\mathcal{S}|}\sum_{X_\ell\in\mathcal{S}}I(X_j,X_\ell).
\label{eq:mrmr_mid}
\end{equation}
Here, $I$ denotes mutual information. The first feature maximized relevance alone because no features had yet been selected. Subsequent features were added greedily by maximizing the MID score among the remaining candidates. Selection continued to produce an ordered ranking of all 90 features within each training partition. A feature's rank indicated its position in this relevance--redundancy selection sequence rather than a statistical significance level or an effect size.

During inner cross-validation, mRMR-MID was fitted using only the inner-training participants and labels. After selecting the feature count for a classifier, the ranking was recomputed using all participants in the corresponding outer-training partition. The classifier was then fitted using the selected number of features from that ranking. Thus, both feature identities and feature counts could vary between outer folds.

For descriptive feature reporting, the position of each feature was averaged across the five outer-training rankings for a task. Features were ordered by increasing mean position, with ties resolved by feature name. The ten earliest features formed that task's aggregate top-ten list. These lists summarized selection priority across training partitions and were not used as fixed input sets during cross-validation.

\subsubsection{Classifiers and Cross-Validation}

The evaluated classifiers were Random Forest, LightGBM, CatBoost, XGBoost, and linear support vector machine. These classifiers provided linear and tree-ensemble approaches to relating the selected measurements to frailty categories. The linear support vector machine represented a linear decision boundary in the selected feature space, while the tree-based models allowed nonlinear decision rules and combinations of feature values. All reported classifiers were unweighted. Each classifier was evaluated using the first $K$ features from the mRMR-MID ranking fitted on the applicable training partition, with $K\in\{10,20,\ldots,90\}$. This grid examined candidate representations ranging from ten features to the complete catalogue.

The classifier-specific results reported here used nested stratified cross-validation with five outer folds and three inner folds at the participant level. Predefined outer partitions were retained across tasks, with each binary task restricted to participants in its two reference categories. Within each outer-training partition, three-fold stratified inner cross-validation evaluated the candidate feature counts separately for each classifier. On each inner split, mRMR-MID was fitted using only the inner-training participants and labels. The classifier was fitted using the resulting top-$K$ features and evaluated on the inner-validation participants.

Stratification approximately preserved class proportions within the applicable partitions. Because each participant contributed a single feature vector, frames from the same participant could not be divided between training and evaluation partitions. Outer-test participants were excluded from feature ranking, feature-count selection, and model fitting. Each classifier was evaluated on them only after its inner selection and outer-training refit were complete.

\subsubsection{Performance Metrics and Model Selection}

The selection metric was micro-averaged area under the receiver operating characteristic curve, denoted by $\mu$AUC. ROC-AUC summarized discrimination across decision thresholds. In the three-class task, micro-averaging aggregated the class-indicator decisions into a single discrimination measure. This provided an overall score for comparing candidate feature counts within a classifier and task rather than a separate estimate for each class.

For each classifier and outer fold, $K$ was selected by maximizing mean inner-validation $\mu$AUC across the three inner folds. Ties were resolved in favor of the smaller feature count. The mRMR-MID ranking was then recomputed using the complete outer-training partition, and the classifier was fitted using its selected top-$K$ features. This fitted model was evaluated once on the held-out outer-test partition. Repeating the procedure across five outer folds yielded five test evaluations and five selected feature counts for each classifier and task. Performance was reported as the mean and standard deviation of these outer-test metrics, and selected feature counts were listed in outer-fold order.

The performance table presents separate nested evaluations of all five classifiers. Rows were ordered descriptively by mean outer-test $\mu$AUC. This ordering summarized observed classifier-specific performance and did not represent an additional classifier-selection step inside the reported estimator. In particular, the highest row was not interpreted as an independently validated estimate of a procedure that automatically chooses among classifiers. The classifier with the highest reported mean outer-test $\mu$AUC for each task was illustrated in the post hoc feature analysis. This choice was descriptive and did not change the held-out predictions or feature-count selection within its classifier-specific evaluation.

Cohen's $\kappa$ complemented discrimination by quantifying agreement between predicted and reference categories beyond chance. Quadratic weighting was used for the ordered three-class outcome, assigning a larger penalty to confusion between nonfrail and frail categories than to confusion between adjacent categories. Unweighted Cohen's $\kappa$ was used for the binary tasks. Agreement was calculated from the same outer-test predictions used to evaluate discrimination rather than from a separately optimized configuration. Both metrics were reported as the mean and standard deviation across the five outer-test folds.

\subsection{Model Interpretation and Feature Recurrence}

SHAP analysis characterized feature contributions to fitted model outputs relative to a reference output~\cite{lundberg2017unified}. Explanations were obtained for held-out outer-test participants using models fitted on the corresponding outer-training partitions. Each participant therefore contributed explanations only from a model fitted without that participant's outer-test fold. The displayed classifier-specific analyses comprised CatBoost for three-class classification, linear support vector machine for frail versus nonfrail and frail versus prefrail classification, and Random Forest for prefrail versus nonfrail classification. Each analysis retained the feature counts selected within its five outer folds.

Feature contributions were aligned by their catalogue identifiers before summarization. 
Pooled summaries averaged absolute SHAP values only over attributions that were computed. For each outer fold, explanations used that fold's fitted top-$K$ mRMR feature set; features absent from a participant's fold-specific model were not assigned zero contribution, and that participant--feature pair was omitted. Accordingly, $\mathcal{I}_j$ denotes the held-out participants for whom feature $j$ appeared in the fitted model, and $B_j$ is conditional on fold-specific selection. Because $K$ varied across folds, $n_j$ could be smaller than the total number of held-out participants for a task.

For feature $j$, let $n_j=|\mathcal{I}_j|$. For frail versus nonfrail and frail versus prefrail classification, the explainer returned one class-wise output per participant ($C=1$); pooled importance followed Equation~\ref{eq:shap_binary}. For three-class classification, Equation~\ref{eq:shap_multiclass} applied with $C=3$. For prefrail versus nonfrail classification, the same pooled form as Equation~\ref{eq:shap_multiclass} applied with $C=2$ because the Random Forest explainer returned separate attributions for each class label; this case is not described by Equation~\ref{eq:shap_binary} alone. Tree-based explainers were used for CatBoost and Random Forest (native SHAP for CatBoost), and a linear explainer for the support vector machine, each with a background sample from the corresponding outer-training partition.
\begin{equation}
B_j=\frac{1}{n_j}\sum_{i\in\mathcal{I}_j}|\phi_{ij}|.
\label{eq:shap_binary}
\end{equation}
Here, $\phi_{ij}$ denotes the held-out attribution for feature $j$ and participant $i$ when $C=1$.

\begin{equation}
B_j=\frac{1}{n_j C}\sum_{i\in\mathcal{I}_j}\sum_{c=1}^{C}|\phi_{ijc}|,
\qquad C=3.
\label{eq:shap_multiclass}
\end{equation}
Here, $\phi_{ijc}$ denotes the attribution for feature $j$, participant $i$, and class output $c$.

Larger mean absolute values indicated greater average contribution magnitude. Averaging absolute values prevented positive and negative contributions from canceling but did not preserve the direction of a feature's association with a frailty category. The summaries did not establish a biomechanical mechanism or a between-group difference in feature values. Attribution magnitudes were interpreted within the output scale of the corresponding classifier rather than compared directly across different classifiers.

SHAP explanations used the complete feature subset fitted in each outer fold. Features were ordered by their pooled importance, and the ten highest-importance features were displayed for clarity. This limit applied to visualization rather than model fitting. A pooled top-ten display could include features drawn from different fold-specific input sets and did not imply a fixed ten-feature model. The fold-specific values of $K$ were included in the figure captions.

Feature recurrence was summarized by comparing the top ten entries of the four aggregate mRMR-MID rankings. Frequency was the number of task-specific top-ten lists containing a feature. Features appearing in at least two lists were included in the cross-task summary. Equal frequencies were treated as tied, with alphabetical presentation within frequency groups. Recurrence described shared priorities across classification tasks rather than selection frequency or stability across individual folds.

\section{Results}\label{sec:results}

The gaitomics catalogue comprised 90 handcrafted features, including 72 first-order and 18 second-order descriptors. First-order features summarized silhouette geometry, temporal variation, and gait energy image intensity distributions. Second-order features characterized co-occurrence texture and gradient- or optical-flow-based patterns. Definitions are provided in the Supplementary Material. Analyses included three-class classification of 66 participants and pairwise classification of frail versus nonfrail participants ($n=42$), frail versus prefrail participants ($n=41$), and prefrail versus nonfrail participants ($n=49$).

Each classifier was evaluated separately using nested stratified cross-validation with five outer and three inner folds at the participant level. Feature ranking and feature-count selection were confined to the relevant training partitions. Consequently, the five fitted models contributing to a classifier-specific summary could differ in both their selected feature counts and feature identities. Performance values are mean $\pm$ standard deviation across five held-out outer-test evaluations. Aggregate mRMR-MID rankings and pooled held-out SHAP summaries provide complementary descriptions of feature priority and model contribution.

\subsection{Task-specific feature rankings}\label{sec:mrmr}\label{sec:univariate}

Tables~\ref{tab:mrmr_top10_three_class}--\ref{tab:mrmr_top10_prefrail_vs_nonfrail} summarize feature priority by mean mRMR-MID position across the five outer-training rankings. Lower mean positions indicate earlier selection across these rankings. These summaries describe task-specific feature priority rather than a common feature subset imposed on all folds. The actual input features were determined separately within the nested training partitions.

\subsubsection{Three-class frailty classification}

Mean absolute bilateral silhouette difference ranked first, followed by maximum contour perimeter and temporal standard deviation of foreground-pixel area (Table~\ref{tab:mrmr_top10_three_class}). Mean signed horizontal optical flow ranked fourth, followed by mean solidity, the vertical-to-horizontal optical-flow ratio, and the mean Sobel-gradient magnitude of consecutive-frame silhouette differences. The remaining features described the temporal mean and standard deviation of log-transformed Hu descriptor indexed 1 and gait energy image intensity variability. Seven of the ten features were first-order and three were second-order. Thus, the aggregate ranking combined silhouette geometry and temporal variation with motion-related measurements.

\begin{table}[htbp]
\centering
\caption{Ten highest-priority features for three-class frailty classification among 66 participants. Features are ordered by their mean position across five mRMR-MID rankings, each computed using only the corresponding outer-training participants and labels. Ties are resolved by feature name. These aggregate rankings summarize feature priority across training partitions and do not represent a single input set shared by all fitted models. Feature-construction categories follow the Methods.}
\label{tab:mrmr_top10_three_class}
\footnotesize
\setlength{\tabcolsep}{3pt}
\renewcommand{\arraystretch}{1.2}
\begin{tabular}{@{}c >{\raggedright\arraybackslash}p{0.28\textwidth} ll >{\raggedright\arraybackslash}p{0.30\textwidth}@{}}
\toprule
\textbf{Rank} & \textbf{Feature} & \textbf{Family} & \textbf{Order} & \textbf{Measure} \\
\midrule
1 & \texttt{temp\_\allowbreak sym\_\allowbreak absdiff\_\allowbreak mean} & Temporal & First-order & Mean absolute bilateral silhouette difference \\
2 & \texttt{shape\_\allowbreak peri\_\allowbreak max} & Shape & First-order & Maximum contour perimeter over frames \\
3 & \texttt{temp\_\allowbreak area\_\allowbreak std} & Temporal & First-order & Temporal standard deviation of foreground-pixel area \\
4 & \texttt{st\_\allowbreak flow\_\allowbreak fx\_\allowbreak mean} & Spatiotemporal & Second-order & Mean signed horizontal optical-flow component \\
5 & \texttt{shape\_\allowbreak solidity\_\allowbreak mean} & Shape & First-order & Temporal mean of contour solidity \\
6 & \texttt{st\_\allowbreak flow\_\allowbreak vert\_\allowbreak ratio\_\allowbreak mean} & Spatiotemporal & Second-order & Mean vertical-to-horizontal optical-flow ratio \\
7 & \texttt{st\_\allowbreak fdt\_\allowbreak sobel\_\allowbreak mean} & Spatiotemporal & Second-order & Mean spatial Sobel-gradient magnitude of consecutive-frame silhouette differences \\
8 & \texttt{shape\_\allowbreak hu1\_\allowbreak mean} & Shape & First-order & Temporal mean of log-transformed Hu descriptor indexed 1 \\
9 & \texttt{gei\_\allowbreak first\_\allowbreak std} & GEI first-order & First-order & GEI standard deviation \\
10 & \texttt{shape\_\allowbreak hu1\_\allowbreak std} & Shape & First-order & Temporal standard deviation of log-transformed Hu descriptor indexed 1 \\
\bottomrule
\end{tabular}
\end{table}

\subsubsection{Pairwise frailty classification}

Bilateral silhouette difference also ranked first in both pairwise comparisons involving frail participants. For frail versus nonfrail classification, the Sobel-gradient descriptor ranked second, maximum perimeter ranked third, and horizontal optical flow ranked fourth (Table~\ref{tab:mrmr_top10_frail_vs_nonfrail}). Additional features captured aspect-ratio variability, relative foreground-area variation, Hu-moment variability, consecutive-frame difference variability, mean solidity, and contour-area variability. Eight of the ten features were first-order and two were second-order.

For frail versus prefrail classification, horizontal optical flow ranked second, followed by gait energy image texture angular second moment and energy (Table~\ref{tab:mrmr_top10_frail_vs_prefrail}). The remaining features comprised mean Hu descriptor indexed 1, maximum perimeter, mean horizontal and vertical silhouette-centroid positions, the vertical-to-horizontal optical-flow ratio, and aspect-ratio variability. Six of the ten features were first-order and four were second-order. This comparison therefore included both texture and motion descriptors among its highest-priority features.

Prefrail versus nonfrail classification showed a different ranking (Table~\ref{tab:mrmr_top10_prefrail_vs_nonfrail}). Maximum contour area ranked first, followed by foreground-area variability and minimum Hu descriptor indexed 6. Motion descriptors included the vertical-to-horizontal optical-flow ratio, horizontal optical flow, optical-flow magnitude variability, and both the temporal variability and mean magnitude of the Sobel-gradient descriptor. Mean Hu descriptor indexed 6 and mean solidity completed the list. Five features were first-order and five were second-order. Bilateral silhouette difference was absent from this aggregate top-ten list.

\begin{table}[htbp]
\centering
\caption{Ten highest-priority features for frail versus nonfrail classification among 42 participants. Features are ordered by mean position across five mRMR-MID rankings computed using only the corresponding outer-training participants and labels. Ties are resolved by feature name. The list summarizes training-partition rankings rather than a fixed input set used in every fold.}
\label{tab:mrmr_top10_frail_vs_nonfrail}
\footnotesize
\setlength{\tabcolsep}{3pt}
\renewcommand{\arraystretch}{1.2}
\begin{tabular}{@{}c >{\raggedright\arraybackslash}p{0.28\textwidth} ll >{\raggedright\arraybackslash}p{0.30\textwidth}@{}}
\toprule
\textbf{Rank} & \textbf{Feature} & \textbf{Family} & \textbf{Order} & \textbf{Measure} \\
\midrule
1 & \texttt{temp\_\allowbreak sym\_\allowbreak absdiff\_\allowbreak mean} & Temporal & First-order & Mean absolute bilateral silhouette difference \\
2 & \texttt{st\_\allowbreak fdt\_\allowbreak sobel\_\allowbreak mean} & Spatiotemporal & Second-order & Mean spatial Sobel-gradient magnitude of consecutive-frame silhouette differences \\
3 & \texttt{shape\_\allowbreak peri\_\allowbreak max} & Shape & First-order & Maximum contour perimeter over frames \\
4 & \texttt{st\_\allowbreak flow\_\allowbreak fx\_\allowbreak mean} & Spatiotemporal & Second-order & Mean signed horizontal optical-flow component \\
5 & \texttt{shape\_\allowbreak aspect\_\allowbreak std} & Shape & First-order & Temporal standard deviation of silhouette aspect ratio \\
6 & \texttt{temp\_\allowbreak area\_\allowbreak cv} & Temporal & First-order & Relative temporal variation in foreground-pixel area \\
7 & \texttt{shape\_\allowbreak hu1\_\allowbreak std} & Shape & First-order & Temporal standard deviation of log-transformed Hu descriptor indexed 1 \\
8 & \texttt{st\_\allowbreak fdiff\_\allowbreak std} & Spatiotemporal & First-order & Standard deviation of consecutive-frame silhouette difference \\
9 & \texttt{shape\_\allowbreak solidity\_\allowbreak mean} & Shape & First-order & Temporal mean of contour solidity \\
10 & \texttt{shape\_\allowbreak area\_\allowbreak std} & Shape & First-order & Temporal standard deviation of contour-based silhouette area \\
\bottomrule
\end{tabular}
\end{table}

\begin{table}[htbp]
\centering
\caption{Ten highest-priority features for frail versus prefrail classification among 41 participants. Features are ordered by mean position across five mRMR-MID rankings computed using only the corresponding outer-training participants and labels. Ties are resolved by feature name. GEI denotes gait energy image, GLCM denotes gray-level co-occurrence matrix, and ASM denotes angular second moment.}
\label{tab:mrmr_top10_frail_vs_prefrail}
\footnotesize
\setlength{\tabcolsep}{3pt}
\renewcommand{\arraystretch}{1.2}
\begin{tabular}{@{}c >{\raggedright\arraybackslash}p{0.28\textwidth} ll >{\raggedright\arraybackslash}p{0.30\textwidth}@{}}
\toprule
\textbf{Rank} & \textbf{Feature} & \textbf{Family} & \textbf{Order} & \textbf{Measure} \\
\midrule
1 & \texttt{temp\_\allowbreak sym\_\allowbreak absdiff\_\allowbreak mean} & Temporal & First-order & Mean absolute bilateral silhouette difference \\
2 & \texttt{st\_\allowbreak flow\_\allowbreak fx\_\allowbreak mean} & Spatiotemporal & Second-order & Mean signed horizontal optical-flow component \\
3 & \texttt{gei\_\allowbreak tex\_\allowbreak ASM\_\allowbreak mean} & GEI texture & Second-order & GLCM angular second moment (mean across orientations) \\
4 & \texttt{gei\_\allowbreak tex\_\allowbreak energy\_\allowbreak mean} & GEI texture & Second-order & GLCM energy (mean across orientations) \\
5 & \texttt{shape\_\allowbreak hu1\_\allowbreak mean} & Shape & First-order & Temporal mean of log-transformed Hu descriptor indexed 1 \\
6 & \texttt{shape\_\allowbreak peri\_\allowbreak max} & Shape & First-order & Maximum contour perimeter over frames \\
7 & \texttt{temp\_\allowbreak com\_\allowbreak x\_\allowbreak mean} & Temporal & First-order & Mean horizontal silhouette-centroid position \\
8 & \texttt{temp\_\allowbreak com\_\allowbreak y\_\allowbreak mean} & Temporal & First-order & Mean vertical silhouette-centroid position \\
9 & \texttt{st\_\allowbreak flow\_\allowbreak vert\_\allowbreak ratio\_\allowbreak mean} & Spatiotemporal & Second-order & Mean vertical-to-horizontal optical-flow ratio \\
10 & \texttt{shape\_\allowbreak aspect\_\allowbreak std} & Shape & First-order & Temporal standard deviation of silhouette aspect ratio \\
\bottomrule
\end{tabular}
\end{table}

\begin{table}[htbp]
\centering
\caption{Ten highest-priority features for prefrail versus nonfrail classification among 49 participants. Features are ordered by mean position across five mRMR-MID rankings computed using only the corresponding outer-training participants and labels. Ties are resolved by feature name. The list summarizes training-partition rankings rather than a fixed input set used in every fold.}
\label{tab:mrmr_top10_prefrail_vs_nonfrail}
\footnotesize
\setlength{\tabcolsep}{3pt}
\renewcommand{\arraystretch}{1.2}
\begin{tabular}{@{}c >{\raggedright\arraybackslash}p{0.28\textwidth} ll >{\raggedright\arraybackslash}p{0.30\textwidth}@{}}
\toprule
\textbf{Rank} & \textbf{Feature} & \textbf{Family} & \textbf{Order} & \textbf{Measure} \\
\midrule
1 & \texttt{shape\_\allowbreak area\_\allowbreak max} & Shape & First-order & Maximum contour-based silhouette area over frames \\
2 & \texttt{temp\_\allowbreak area\_\allowbreak std} & Temporal & First-order & Temporal standard deviation of foreground-pixel area \\
3 & \texttt{shape\_\allowbreak hu6\_\allowbreak min} & Shape & First-order & Minimum log-transformed Hu descriptor indexed 6 over frames \\
4 & \texttt{st\_\allowbreak flow\_\allowbreak vert\_\allowbreak ratio\_\allowbreak mean} & Spatiotemporal & Second-order & Mean vertical-to-horizontal optical-flow ratio \\
5 & \texttt{shape\_\allowbreak hu6\_\allowbreak mean} & Shape & First-order & Temporal mean of log-transformed Hu descriptor indexed 6 \\
6 & \texttt{st\_\allowbreak flow\_\allowbreak fx\_\allowbreak mean} & Spatiotemporal & Second-order & Mean signed horizontal optical-flow component \\
7 & \texttt{st\_\allowbreak flow\_\allowbreak mag\_\allowbreak std} & Spatiotemporal & Second-order & Standard deviation of optical-flow magnitude \\
8 & \texttt{st\_\allowbreak fdt\_\allowbreak sobel\_\allowbreak std\_\allowbreak t} & Spatiotemporal & Second-order & Temporal standard deviation of spatial Sobel-gradient magnitude of consecutive-frame differences \\
9 & \texttt{shape\_\allowbreak solidity\_\allowbreak mean} & Shape & First-order & Temporal mean of contour solidity \\
10 & \texttt{st\_\allowbreak fdt\_\allowbreak sobel\_\allowbreak mean} & Spatiotemporal & Second-order & Mean spatial Sobel-gradient magnitude of consecutive-frame silhouette differences \\
\bottomrule
\end{tabular}
\end{table}

\subsection{Classification performance}\label{sec:classification}

Table~\ref{tab:q3_ml_all_comparisons_five_models} reports classifier-specific outer-test performance and the five feature counts selected by inner cross-validation. Micro-averaged ROC-AUC ($\mu$AUC) summarizes discrimination across decision thresholds, while Cohen's $\kappa$ summarizes agreement between predicted and reference categories beyond chance. Agreement used quadratic weighting for three-class classification and no weighting for the binary tasks. Both metrics describe the same outer-test predictions. Comparisons among classifiers are descriptive and do not establish statistically significant performance differences or validate an automatic procedure for choosing a classifier from its outer-test results.

\subsubsection{Three-class frailty classification}

CatBoost had the highest mean $\mu$AUC for three-class classification ($0.647\pm0.085$), with $\kappa=0.265\pm0.117$. Its selected feature counts were 10, 20, 40, 30, and 10 across outer folds one through five. Linear SVM followed with $\mu$AUC $0.640\pm0.119$ and the highest mean agreement ($\kappa=0.355\pm0.297$), using ten features in four folds and 20 in the remaining fold. Random Forest, XGBoost, and LightGBM achieved mean $\mu$AUC values of $0.611\pm0.059$, $0.591\pm0.099$, and $0.548\pm0.055$, with corresponding agreement values of $0.135\pm0.283$, $0.193\pm0.294$, and $0.035\pm0.153$. Feature counts varied across folds for all classifiers except LightGBM, which selected ten features in every fold. The highest mean discrimination and highest mean agreement were therefore obtained by different classifiers.

\subsubsection{Pairwise frailty classification}

Frail versus nonfrail classification yielded the highest leading mean $\mu$AUC among the pairwise tasks. Linear SVM achieved $\mu$AUC $0.833\pm0.249$ and the highest mean agreement for this comparison ($\kappa=0.608\pm0.363$). It selected 10, 80, 10, 10, and 10 features across the five outer folds. Random Forest followed with $\mu$AUC $0.740\pm0.238$ and $\kappa=0.178\pm0.347$. XGBoost and CatBoost achieved $\mu$AUC values of $0.683\pm0.307$ and $0.670\pm0.187$, with agreement values of $0.171\pm0.436$ and $0.064\pm0.180$. LightGBM yielded $\mu$AUC $0.500\pm0.000$ and $\kappa=0.000\pm0.000$.

For frail versus prefrail classification, linear SVM had both the highest mean discrimination ($\mu$AUC $0.765\pm0.168$) and the highest mean agreement ($\kappa=0.353\pm0.426$). Its selected feature counts were 10, 60, 10, 10, and 40. CatBoost and XGBoost each had reported $\mu$AUC $0.627\pm0.220$, with agreement values of $0.057\pm0.193$ and $0.159\pm0.212$, respectively. Random Forest achieved $\mu$AUC $0.573\pm0.307$ and $\kappa=0.163\pm0.380$. LightGBM again yielded $\mu$AUC $0.500\pm0.000$ and $\kappa=0.000\pm0.000$.

Prefrail versus nonfrail classification showed the weakest discrimination. Random Forest had the highest mean $\mu$AUC ($0.554\pm0.129$), with $\kappa=0.009\pm0.126$, using 40, 30, 80, 20, and 10 features across outer folds. CatBoost achieved $\mu$AUC $0.540\pm0.148$ and $\kappa=0.010\pm0.301$. XGBoost achieved $\mu$AUC $0.540\pm0.179$ and the highest mean agreement for this comparison ($\kappa=0.036\pm0.391$). LightGBM and linear SVM yielded $\mu$AUC values of $0.485\pm0.034$ and $0.402\pm0.183$, with agreement values of $-0.020\pm0.044$ and $-0.011\pm0.308$. Thus, even the leading classifier-specific mean AUC was close to chance, and mean agreement remained close to zero.

The leading classifier-specific summaries differed across tasks. CatBoost led three-class discrimination, linear SVM led both comparisons involving frail participants, and Random Forest led prefrail versus nonfrail discrimination. The selected feature counts also varied between outer folds within each classifier, so these summaries do not identify a single fixed feature count for a task. Several estimates showed substantial variation across outer folds, particularly in the binary comparisons.

\begin{table}[htbp]
\centering
\caption{Classifier-specific nested cross-validation using five outer and three inner stratified folds at the participant level. Within each classifier and outer fold, $K$ maximized mean inner-validation $\mu$AUC over $K\in\{10,20,\ldots,90\}$, with ties resolved by smaller $K$. mRMR-MID was fitted on inner-training partitions for selection and refitted on the complete outer-training partition before final model fitting. The feature-count column lists $K$ for outer folds one through five. Metrics are mean $\pm$ standard deviation across five held-out outer-test folds. Rows are ordered descriptively by mean outer-test $\mu$AUC, with no additional distinction implied for equal displayed values. All classifiers were unweighted. F denotes frail, PF denotes prefrail, and NF denotes nonfrail. Cohen's $\kappa$ used quadratic weighting for three-class classification and no weighting for binary classification.}
\label{tab:q3_ml_all_comparisons_five_models}
\footnotesize
\setlength{\tabcolsep}{3pt}
\renewcommand{\arraystretch}{1.12}
\begin{tabular}{@{}lclccc@{}}
\toprule
\textbf{Comparison} & \textbf{Rank} & Model & $K$ by outer fold & $\mu$AUC & $\kappa$ \\
\midrule
3-class (F/PF/NF) ($n=66$) & 1 & CatBoost & \scriptsize 10{,} 20{,} 40{,} 30{,} 10 & $0.647 \pm 0.085$ & $0.265 \pm 0.117$ \\
 & 2 & Linear SVM & \scriptsize 10{,} 10{,} 10{,} 10{,} 20 & $0.640 \pm 0.119$ & $0.355 \pm 0.297$ \\
 & 3 & Random Forest & \scriptsize 30{,} 10{,} 10{,} 60{,} 10 & $0.611 \pm 0.059$ & $0.135 \pm 0.283$ \\
 & 4 & XGBoost & \scriptsize 40{,} 10{,} 20{,} 50{,} 10 & $0.591 \pm 0.099$ & $0.193 \pm 0.294$ \\
 & 5 & LightGBM & \scriptsize 10{,} 10{,} 10{,} 10{,} 10 & $0.548 \pm 0.055$ & $0.035 \pm 0.153$ \\
\addlinespace
\midrule
F vs NF ($n=42$) & 1 & Linear SVM & \scriptsize 10{,} 80{,} 10{,} 10{,} 10 & $0.833 \pm 0.249$ & $0.608 \pm 0.363$ \\
 & 2 & Random Forest & \scriptsize 10{,} 60{,} 10{,} 20{,} 20 & $0.740 \pm 0.238$ & $0.178 \pm 0.347$ \\
 & 3 & XGBoost & \scriptsize 10{,} 10{,} 20{,} 30{,} 70 & $0.683 \pm 0.307$ & $0.171 \pm 0.436$ \\
 & 4 & CatBoost & \scriptsize 10{,} 10{,} 10{,} 60{,} 30 & $0.670 \pm 0.187$ & $0.064 \pm 0.180$ \\
 & 5 & LightGBM & \scriptsize 10{,} 10{,} 10{,} 10{,} 10 & $0.500 \pm 0.000$ & $0.000 \pm 0.000$ \\
\addlinespace
\midrule
F vs PF ($n=41$) & 1 & Linear SVM & \scriptsize 10{,} 60{,} 10{,} 10{,} 40 & $0.765 \pm 0.168$ & $0.353 \pm 0.426$ \\
 & 2 & CatBoost & \scriptsize 70{,} 30{,} 10{,} 10{,} 20 & $0.627 \pm 0.220$ & $0.057 \pm 0.193$ \\
 & 3 & XGBoost & \scriptsize 10{,} 10{,} 90{,} 40{,} 50 & $0.627 \pm 0.220$ & $0.159 \pm 0.212$ \\
 & 4 & Random Forest & \scriptsize 10{,} 10{,} 10{,} 40{,} 90 & $0.573 \pm 0.307$ & $0.163 \pm 0.380$ \\
 & 5 & LightGBM & \scriptsize 10{,} 10{,} 10{,} 10{,} 10 & $0.500 \pm 0.000$ & $0.000 \pm 0.000$ \\
\addlinespace
\midrule
PF vs NF ($n=49$) & 1 & Random Forest & \scriptsize 40{,} 30{,} 80{,} 20{,} 10 & $0.554 \pm 0.129$ & $0.009 \pm 0.126$ \\
 & 2 & CatBoost & \scriptsize 40{,} 10{,} 30{,} 30{,} 10 & $0.540 \pm 0.148$ & $0.010 \pm 0.301$ \\
 & 3 & XGBoost & \scriptsize 90{,} 20{,} 70{,} 20{,} 10 & $0.540 \pm 0.179$ & $0.036 \pm 0.391$ \\
 & 4 & LightGBM & \scriptsize 10{,} 10{,} 10{,} 10{,} 10 & $0.485 \pm 0.034$ & $-0.020 \pm 0.044$ \\
 & 5 & Linear SVM & \scriptsize 80{,} 70{,} 20{,} 50{,} 90 & $0.402 \pm 0.183$ & $-0.011 \pm 0.308$ \\
\bottomrule
\end{tabular}
\end{table}

\subsection{Post hoc interpretation of the selected models}\label{sec:shap}

Figures~\ref{fig:shap_three_class} and~\ref{fig:shap_pairwise} show the ten highest pooled mean absolute SHAP contributions for the classifier with the largest reported mean outer-test $\mu$AUC in each task. These are descriptive classifier-specific explanations. Each participant was explained using a model fitted on the corresponding outer-training partition, with the full feature subset selected for that fold. The ten-feature display limit therefore does not imply that all fitted models had ten inputs or shared the displayed feature identities. Larger values indicate larger average contribution magnitudes rather than a direction of association with a frailty category. Absolute magnitudes are interpreted within the output scale of the corresponding classifier.

\subsubsection{Three-class classification}

\begin{figure}[htbp]
\centering
\includegraphics[width=0.70\linewidth]{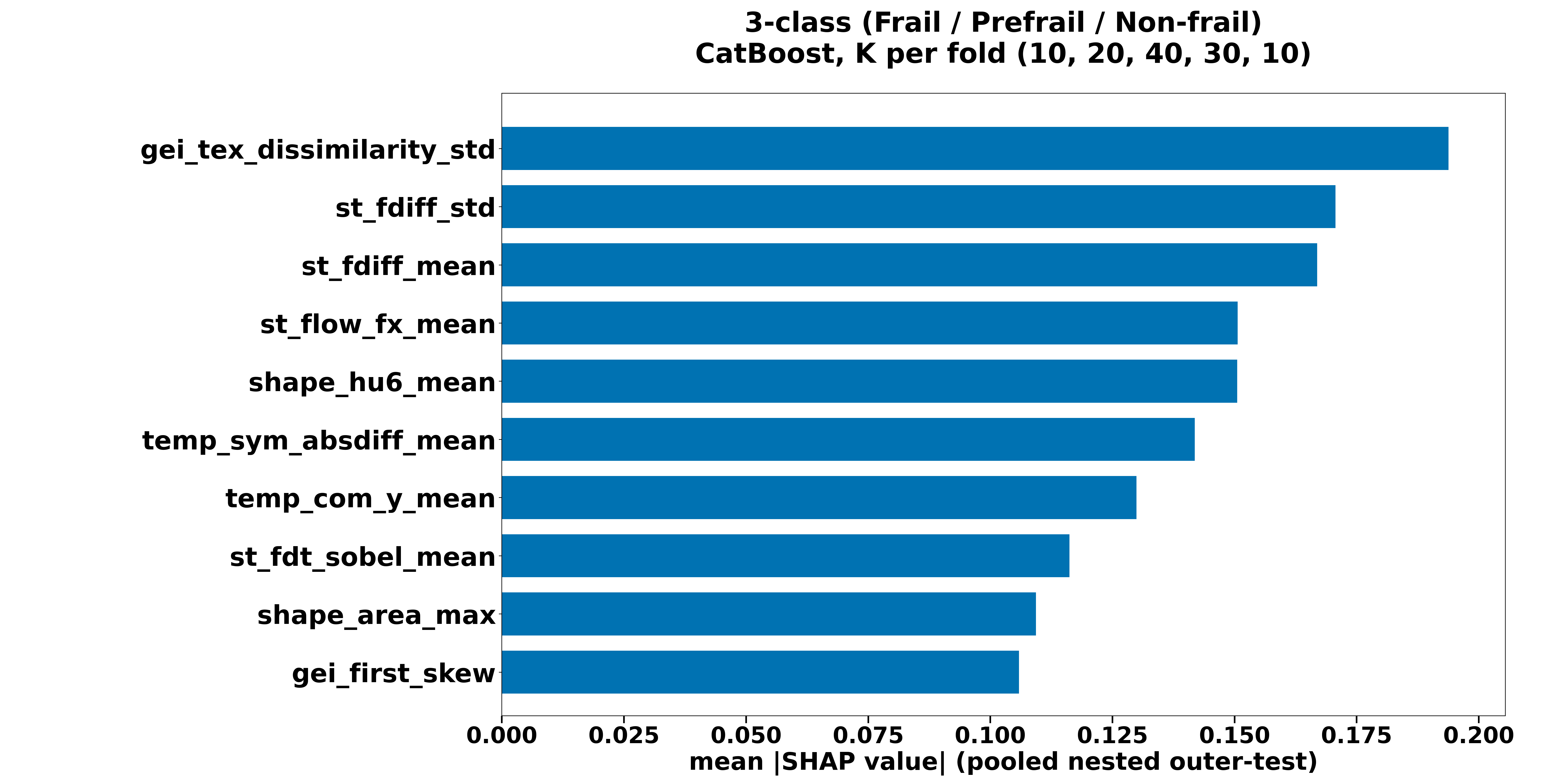}
\caption{SHAP feature importance for the classifier-specific three-class CatBoost analysis. Inner-validation selection yielded 10, 20, 40, 30, and 10 features across outer folds one through five. The corresponding mean outer-test $\mu$AUC was $0.647\pm0.085$. Bars show pooled mean absolute contributions from held-out outer-test participants explained using their corresponding outer-training models. The ten highest-importance features are displayed, while explanations retain each fold's complete fitted input set. Larger values indicate greater average contribution magnitude and do not indicate contribution direction.}
\label{fig:shap_three_class}
\end{figure}

In the CatBoost analysis, variability across orientations in GEI texture dissimilarity had the highest pooled mean absolute contribution (Figure~\ref{fig:shap_three_class}). Variability and mean magnitude of consecutive-frame silhouette differences ranked second and third. Horizontal optical flow ranked fourth, followed by mean Hu descriptor indexed six and bilateral silhouette difference. Mean vertical silhouette-centroid position, mean Sobel-gradient magnitude, maximum contour area, and GEI intensity skewness completed the displayed list. Thus, the leading contributions combined GEI texture, sequence-level image changes, motion, and shape descriptors.

The SHAP display differed from the aggregate mRMR-MID top ten. Bilateral silhouette difference ranked first by mean mRMR position but sixth in the attribution display, while GEI texture dissimilarity variability led the attribution display without appearing in the aggregate mRMR top ten. The two summaries therefore characterized different aspects of the analysis. Aggregate mRMR priority reflects relevance balanced against redundancy across outer-training rankings, whereas the attribution display reflects contributions within fold-specific fitted models. The aggregate top-ten list is not a complete inventory of the inputs available across all folds.

\subsubsection{Pairwise classification}

\begin{figure}[htbp]
\centering
\includegraphics[width=0.80\linewidth]{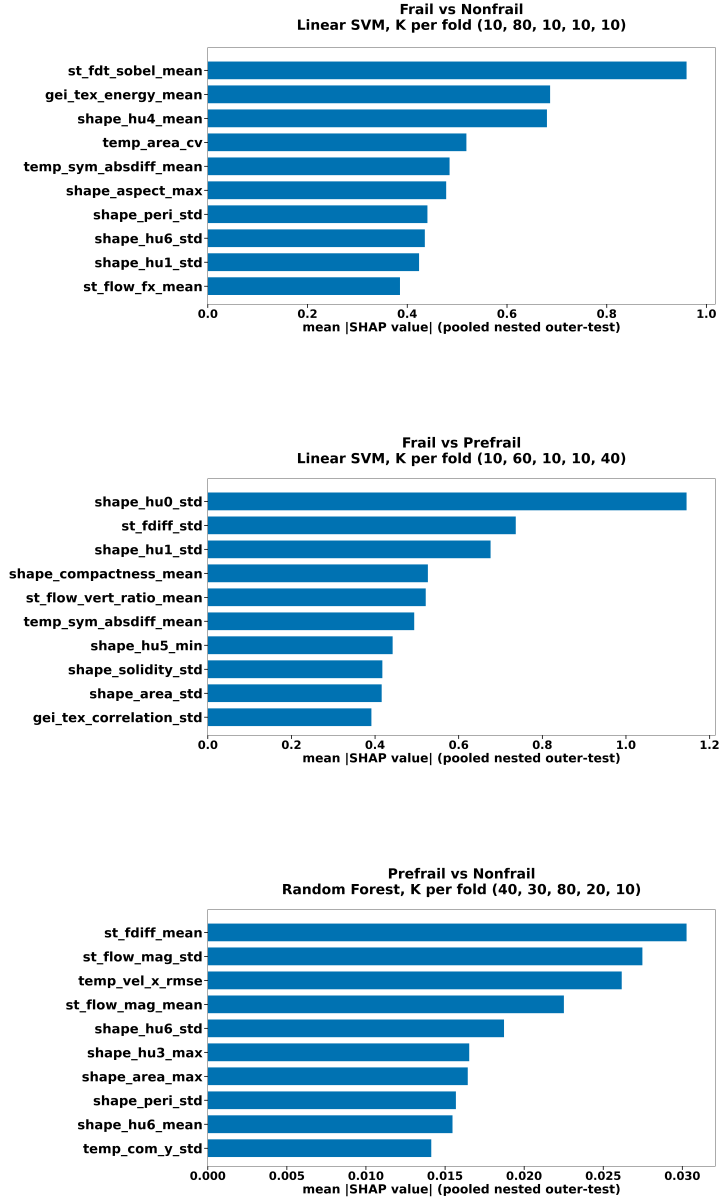}
\caption{SHAP feature importance for the leading classifier-specific pairwise summaries. The upper panel shows frail versus nonfrail linear SVM, with outer-fold feature counts of 10, 80, 10, 10, and 10 and mean outer-test $\mu$AUC $0.833\pm0.249$. The middle panel shows frail versus prefrail linear SVM, with feature counts of 10, 60, 10, 10, and 40 and $\mu$AUC $0.765\pm0.168$. The lower panel shows prefrail versus nonfrail Random Forest, with feature counts of 40, 30, 80, 20, and 10 and $\mu$AUC $0.554\pm0.129$. Bars show pooled mean absolute contributions from held-out outer-test participants. Each panel displays ten features, while explanations use each fold's complete fitted input set. Panels have separate horizontal scales, and magnitudes should be interpreted within the corresponding model output scale.}
\label{fig:shap_pairwise}\label{fig:pipeline}
\end{figure}

For frail versus nonfrail classification, the linear SVM analysis assigned the highest pooled importance to mean Sobel-gradient magnitude, followed by mean GEI texture energy and the mean Hu descriptor indexed four (Figure~\ref{fig:shap_pairwise}). Relative foreground-area variability ranked fourth and bilateral silhouette difference ranked fifth. Maximum aspect ratio, perimeter variability, variability in Hu descriptors indexed six and one, and horizontal optical flow completed the top ten. The displayed contributions therefore included gradient and texture information alongside silhouette geometry and temporal variability.

For frail versus prefrail classification, the linear SVM analysis was led by variability in the Hu descriptor indexed zero, consecutive-frame difference variability, and variability in the Hu descriptor indexed one. Mean compactness ranked fourth, followed by the vertical-to-horizontal optical-flow ratio and bilateral silhouette difference. Minimum Hu descriptor indexed five, solidity variability, contour-area variability, and variability across orientations in GEI texture correlation completed the displayed list. Bilateral silhouette difference remained among the ten largest contributions but was not the leading feature.

For prefrail versus nonfrail classification, mean consecutive-frame silhouette difference had the largest pooled importance in the Random Forest analysis. Optical-flow magnitude variability ranked second, followed by the horizontal sequence-velocity descriptor \texttt{temp\_vel\_x\_rmse} and mean optical-flow magnitude. Variability in Hu descriptor indexed six, maximum Hu descriptor indexed three, maximum contour area, perimeter variability, mean Hu descriptor indexed six, and vertical silhouette-centroid variability completed the top ten. Bilateral silhouette difference was absent from the displayed list. This importance profile describes the inputs contributing to the fitted models despite the weak discrimination and near-zero mean agreement observed for this task.

\subsection{Feature recurrence across classification tasks}\label{sec:feature_recurrence}

Ten features appeared in the aggregate top-ten mRMR-MID lists of at least two classification tasks (Table~\ref{tab:mrmr_overall_top10_recurrence}). Mean signed horizontal optical flow appeared in all four lists. Five features appeared in three lists each. Bilateral silhouette difference and maximum perimeter recurred in the three-class task and both comparisons involving frail participants. Mean solidity and mean Sobel-gradient magnitude recurred in the three-class, frail versus nonfrail, and prefrail versus nonfrail tasks. The vertical-to-horizontal optical-flow ratio recurred in the three-class, frail versus prefrail, and prefrail versus nonfrail tasks.

Four additional features appeared in two lists each. These comprised aspect-ratio variability, the temporal mean and standard deviation of Hu descriptor indexed 1, and foreground-area variability. Seven recurrent features were first-order and three were second-order. Recurrence therefore involved both direct silhouette measurements and motion-related descriptors. No gait energy image descriptor appeared in more than one aggregate top-ten list.

Bilateral silhouette difference retained the highest aggregate mRMR priority for three-class classification and both comparisons involving frail participants. It also appeared in the corresponding SHAP top-ten displays, ranking sixth, fifth, and sixth, respectively. The leading attribution features differed across these analyses, comprising GEI texture dissimilarity variability for three-class classification, mean Sobel-gradient magnitude for frail versus nonfrail classification, and Hu-descriptor variability for frail versus prefrail classification. Prefrail versus nonfrail classification combined an aggregate ranking led by contour area with an attribution display led by frame differences and optical-flow descriptors. These summaries indicate shared selection priorities alongside task- and model-dependent attribution profiles. Recurrence counts membership in the four aggregate lists and does not establish fold-wise selection stability or an incremental predictive benefit.

\begin{table}[htbp]
\centering
\caption{Features appearing in the aggregate top-ten mRMR-MID lists of at least two tasks. Each task-specific list is based on mean feature position across five outer-training rankings. Frequency counts membership across the four task-specific lists and is not a fold-selection frequency. All ten recurrent features are shown. Alphabetical ordering within a frequency group does not imply an additional importance ranking. F denotes frail, PF denotes prefrail, and NF denotes nonfrail.}
\label{tab:mrmr_overall_top10_recurrence}
\footnotesize
\setlength{\tabcolsep}{3pt}
\renewcommand{\arraystretch}{1.2}
\begin{tabular}{@{}c >{\raggedright\arraybackslash}p{0.27\textwidth} >{\raggedright\arraybackslash}p{0.29\textwidth} >{\raggedright\arraybackslash}p{0.29\textwidth}@{}}
\toprule
\textbf{Frequency} & \textbf{Feature} & \textbf{Measure} & \textbf{Comparisons} \\
\midrule
4/4 & \texttt{st\_\allowbreak flow\_\allowbreak fx\_\allowbreak mean} & Mean signed horizontal optical-flow component & Three-class, F vs NF, F vs PF, PF vs NF \\
3/4 & \texttt{shape\_\allowbreak peri\_\allowbreak max} & Maximum contour perimeter over frames & Three-class, F vs NF, F vs PF \\
3/4 & \texttt{shape\_\allowbreak solidity\_\allowbreak mean} & Temporal mean of contour solidity & Three-class, F vs NF, PF vs NF \\
3/4 & \texttt{st\_\allowbreak fdt\_\allowbreak sobel\_\allowbreak mean} & Mean spatial Sobel-gradient magnitude of consecutive-frame silhouette differences & Three-class, F vs NF, PF vs NF \\
3/4 & \texttt{st\_\allowbreak flow\_\allowbreak vert\_\allowbreak ratio\_\allowbreak mean} & Mean vertical-to-horizontal optical-flow ratio & Three-class, F vs PF, PF vs NF \\
3/4 & \texttt{temp\_\allowbreak sym\_\allowbreak absdiff\_\allowbreak mean} & Mean absolute bilateral silhouette difference & Three-class, F vs NF, F vs PF \\
2/4 & \texttt{shape\_\allowbreak aspect\_\allowbreak std} & Temporal standard deviation of silhouette aspect ratio & F vs NF, F vs PF \\
2/4 & \texttt{shape\_\allowbreak hu1\_\allowbreak mean} & Temporal mean of log-transformed Hu descriptor indexed 1 & Three-class, F vs PF \\
2/4 & \texttt{shape\_\allowbreak hu1\_\allowbreak std} & Temporal standard deviation of log-transformed Hu descriptor indexed 1 & Three-class, F vs NF \\
2/4 & \texttt{temp\_\allowbreak area\_\allowbreak std} & Temporal standard deviation of foreground-pixel area & Three-class, PF vs NF \\
\bottomrule
\end{tabular}
\end{table}
\section{Discussion}\label{sec12}

This study examined whether a predefined catalog of silhouette-derived gait measurements could support interpretable frailty classification. Training-partition-specific mRMR-MID selection, nested cross-validation, and held-out SHAP analysis provided complementary descriptions of feature priority, predictive performance, and model behavior. Bilateral silhouette configuration was repeatedly prioritized during feature selection, while texture, consecutive-frame differences, shape, and motion descriptors contributed differently across the explained models. Classification performance also depended on the comparison, with the highest mean pairwise discrimination between frail and nonfrail participants and limited separation of prefrail and nonfrail participants. The central contribution is a framework for connecting predefined gait measurements with model predictions while examining the extent and limitations of their predictive information.

The distinction between first-order and second-order features helps explain what the catalog contributes. Under the adopted taxonomy, the 72 first-order descriptors summarize silhouette geometry, temporal changes in individual measurements, GEI intensity distributions, and aggregate frame differences. Area, perimeter, aspect ratio, solidity, and Hu moments characterize projected body configuration. Their sequence-level summaries describe how that configuration varies during walking. The 18 second-order descriptors represent GEI texture and gradient- or optical-flow-based patterns. These add information about spatial relationships and the distribution or direction of change between frames. Two sequences could have similar overall area variation yet differ in where movement occurs or in its direction, providing a rationale for considering both feature families. GEI texture offers a further view of how accumulated silhouette occupancy is organized spatially, beyond its overall intensity distribution. mRMR-MID provides a practical means of choosing among these related measurements by balancing relevance to frailty labels against redundancy with features already selected~\cite{peng2005feature}. The aggregate three-class top-ten ranking contained seven first-order and three second-order descriptors, combining bilateral configuration, geometry, temporal variation, GEI intensity variability, and motion measurements (Table~\ref{tab:mrmr_top10_three_class}). This list summarizes priority across outer-training rankings rather than a fixed subset shared by every fitted model. In the three-class CatBoost SHAP display, GEI texture dissimilarity variability ranked first, while horizontal optical flow and the Sobel gradient of consecutive-frame differences ranked fourth and eighth. These findings illustrate the contribution of spatial organization and sequence-level changes alongside silhouette geometry. Determining the incremental predictive benefit of either feature order would require a direct ablation comparison.

Bilateral silhouette difference ranked first in the aggregate mRMR-MID lists for three-class classification and both comparisons involving frail participants. It also appeared in the corresponding SHAP top-ten displays, ranking sixth, fifth, and sixth, respectively (Figures~\ref{fig:shap_three_class} and~\ref{fig:shap_pairwise}). Its recurring selection priority and attribution presence direct attention to body configuration during walking, which can be influenced by posture, limb excursion, coordination, and musculoskeletal function~\cite{gao2026gait,pirker2017gait,serra2019muscle}. However, a left--right difference in a projected silhouette is distinct from bilateral step timing, loading, or joint kinematics. The appropriate interpretation is that this image-derived configuration measure was informative during feature selection and contributed to the explained predictions. Establishing its relationship to a specific biomechanical impairment requires synchronized functional or kinematic assessment~\cite{martinez2015frailty}. The importance rankings alone do not establish increased clinical gait asymmetry with frailty.

Area-related descriptors provide a complementary description of silhouette extent and its change over time. Foreground-pixel area variability appeared in the aggregate three-class and prefrail versus nonfrail top-ten rankings, while contour-area variability appeared in the frail versus nonfrail aggregate top ten. The attribution displays emphasized different area summaries. Maximum contour area ranked ninth in three-class classification and seventh in prefrail versus nonfrail classification. Relative foreground-area variability ranked fourth in frail versus nonfrail classification, while contour-area variability ranked ninth in frail versus prefrail classification. Mean solidity recurred during feature selection, and solidity variability appeared in the frail versus prefrail SHAP display. Together, these measurements describe projected silhouette extent, occupancy of its convex hull, and variation across the sampled sequence. Contour area and foreground-pixel area nevertheless remain distinct quantities. Their temporal summaries should be interpreted as image-based dynamics rather than direct measures of stride variability or balance. Reduced visible variation could reflect restricted movement and does not necessarily imply greater stability. The importance plots also do not establish whether any frailty group had greater or smaller feature values.

The performance pattern was consistent with greater mean separation of the extreme frailty categories. Among the classifier-specific summaries, linear SVM achieved the highest pairwise mean micro-AUC for frail versus nonfrail classification ($0.833\pm0.249$) and also led frail versus prefrail classification ($0.765\pm0.168$). Random Forest had the highest mean micro-AUC for prefrail versus nonfrail classification ($0.554\pm0.129$), while CatBoost led three-class discrimination ($0.647\pm0.085$) (Table~\ref{tab:q3_ml_all_comparisons_five_models}). These results describe separate nested evaluations of each classifier, with substantial variability across several outer-fold estimates. Characterizing differences across frailty severity is relevant to gait-based assessment~\cite{jiang2025gait}, but the distinct feature profiles indicate that the comparisons need not depend on the same gait characteristics. Bilateral silhouette difference was absent from the prefrail versus nonfrail aggregate top ten, which combined five first-order and five second-order descriptors. Horizontal optical flow appeared in all four aggregate top-ten lists, indicating a shared selection priority. This recurrence does not establish a universal clinical marker or eliminate task dependence in the remaining measurements.

The limited prefrail versus nonfrail discrimination may reflect the heterogeneity of prefrailty. Individuals can meet different combinations of Fried criteria and have different functional profiles despite sharing the same category. Previous work has described heterogeneity in the manifestations of frailty and the multidimensional nature of prefrailty~\cite{xue2008initial,sezgin2020pre}. Consequently, early differences may overlap with variation among nonfrail individuals or involve aspects of health that a walking silhouette does not capture. This offers a plausible explanation for the weak discrimination and distinct combination of shape, temporal, and motion descriptors, but requires confirmation using individual Fried components and clinical characteristics. The leading classifier's mean agreement was also close to zero ($\kappa=0.009\pm0.126$), limiting claims about reliable categorical separation. Its SHAP profile therefore provides an exploratory description of fitted-model behavior whose clinical relevance requires further validation. Cross-sectional classification also remains distinct from predicting an individual's subsequent progression.

The fold-specific feature counts demonstrate the data dependence of the selected representations. The leading three-class CatBoost configuration used between ten and 40 features across outer folds. Linear SVM selected ten or 80 features for frail versus nonfrail classification and ten, 40, or 60 features for frail versus prefrail classification. The prefrail versus nonfrail Random Forest configuration used between ten and 80 features. A larger subset may allow a classifier to combine information that is less prominent among the earliest-ranked features, but these selections do not establish a biological requirement for an exact number of measurements. Their variation also precludes describing each task using a single fixed feature count. Ten was the smallest subset evaluated and therefore cannot be regarded as the minimum sufficient number. The practical finding is that feature-count selection depended on the training partition, classifier, and target comparison, reinforcing the importance of evaluating the complete selection procedure within nested validation.

SHAP characterized how the explained models used their inputs, complementing the relevance--redundancy criterion used during feature selection~\cite{lundberg2017unified}. For example, bilateral silhouette difference ranked first in the aggregate three-class mRMR list and sixth in its SHAP display. Conversely, GEI texture dissimilarity variability led the three-class attribution display without appearing in its aggregate mRMR top ten. These summaries describe different properties and stages of the analysis. The aggregate mRMR list reflects mean selection position across outer-training partitions, whereas SHAP summarizes contributions within fitted models on held-out participants. Each visualization was capped at the ten largest pooled mean absolute contributions for readability. The displayed CatBoost, linear SVM, and Random Forest analyses retained the complete feature subset fitted in each outer fold. Consequently, ten displayed features did not imply ten fitted inputs or identical feature membership across folds. Mean absolute SHAP summarizes contribution magnitude without preserving the direction of association with frailty or representing a physiological effect size. Agreement between selection priority and attribution within these overlapping cohorts should be interpreted as analytical consistency rather than independent replication.

The distinction between discrimination and categorical agreement is also clinically relevant. In three-class classification, CatBoost had the highest mean micro-AUC ($0.647\pm0.085$), with quadratic-weighted agreement of $\kappa=0.265\pm0.117$. Linear SVM had slightly lower mean discrimination ($0.640\pm0.119$) but higher mean agreement ($\kappa=0.355\pm0.297$). These descriptive differences illustrate why discrimination across thresholds does not ensure the highest categorical agreement at a model's decision rule. They do not establish statistically significant differences between classifiers. Clinical evaluation should therefore examine class-specific errors, calibration, and decision thresholds alongside aggregate discrimination, particularly when the intended use involves identifying prefrail individuals for further assessment.

The framework could support review of a predicted category alongside explicit measurements of the observed gait. Silhouettes retain geometry and movement while removing color and much appearance information, although they do not guarantee anonymity. For earlier-stage assessment, demographic and clinical information and repeated measurements may help address information that is not visible in a single walking sequence. Examination of individual Fried criteria may further clarify which aspects of the reference phenotype are reflected in the gait descriptors~\cite{zhang2025gait,sezgin2020pre}. Such integration should be evaluated against a defined purpose, distinguishing contemporaneous frailty classification from monitoring functional change or predicting future decline. The present results support further investigation of this approach while leaving its clinical utility to be established.

Several limitations warrant consideration. The small, uneven cohort ($n=66$ with 17 frail, 24 prefrail, and 25 nonfrail participants), fold-to-fold variability, and unweighted classifiers limit precision and require class-specific evaluation beyond micro-AUC. Nested training-only ranking and inner-loop feature-count selection reduced selection-related optimism~\cite{varma2006bias}, while classifier comparisons remained descriptive and independent external validation remains necessary. The ten-feature grid, absence of feature-family ablations, and untested selection stability limit conclusions about optimal subset size and incremental information. Walking aids may alter body configuration and obscure parts of the participant even when excluded from the retained mask, potentially confounding frailty associations. Camera geometry, viewing direction, walking distance, clothing, illumination, resolution, segmentation, and ROI processing may also affect measurements and generalizability. Test--retest reliability and biomechanical validity remain unestablished. Finally, overlap with the walking component of the Fried Phenotype, potentially similar changes from other mobility-limiting conditions, and the cross-sectional design limit claims about broader frailty specificity, causation, and future progression.

Future work should combine external and longitudinal validation with direct comparisons of descriptor-based and transformer-based gait representations. Learned representations may capture additional movement patterns, while the current catalog provides an interpretable reference for assessing what they add. Integrating these approaches with clinical information and repeated assessments may be particularly useful for investigating prefrail versus nonfrail overlap. Overall, the findings identify task-dependent selection and attribution profiles spanning silhouette geometry, temporal variation, texture, and derived motion information. These measurements provide candidates for further investigation, with their reliability and predictive value requiring confirmation in larger independent cohorts.

\section{Conclusion}\label{sec13}

The present study establishes silhouette-based gaitomics as an exploratory framework for relating predefined video-derived gait measurements to frailty classification. Bilateral silhouette difference was repeatedly prioritized during feature selection, while texture, consecutive-frame differences, geometry, and motion contributed differently across the explained models. These findings characterize model behavior without establishing specific biomechanical impairments. Mean discrimination was highest between frail and nonfrail participants, while prefrail versus nonfrail classification remained close to chance. The framework supports further investigation of task-specific gait representations and transparent evaluation of their predictive information. Future work will examine whether transformer-based gait representations and integration with demographic and clinical information improve classification. Larger independent cohorts, measurement-reliability assessment, and longitudinal evaluation are needed to establish generalizability and clinical relevance before silhouette-based gaitomics can be considered a useful complement to established frailty assessment.

\bibliography{sn-bibliography}

\end{document}